# Very Short Literature Survey From Supervised Learning To Surrogate Modeling

By: Altay Brusan

**Abstract:** The past century was era of linear systems. Either systems (especially industrial ones) were simple (quasi)linear or linear approximations were accurate enough. In addition, just at the ending decades of the century profusion of computing devices were available, before then due to lack of computational resources it was not easy to evaluate available nonlinear system studies. At the moment both these two conditions changed, systems are highly complex and also pervasive amount of computation strength is cheap and easy to achieve. For recent era, a new branch of supervised learning well known as surrogate modeling (meta-modeling, surface modeling) has been devised which aimed at answering new needs of modeling realm. This short literature survey is on to introduce surrogate modeling to whom is familiar with the concepts of supervised learning. Necessity, challenges and visions of the topic are considered.

## 1. INTRODUCTION

Since the quality of a model typically determines an upper bound on the quality of the final problem solution, modeling is often the bottleneck in the development of the whole system. The process of model construction is modeling (Learning) which is schematically got represented at Fig. 1.

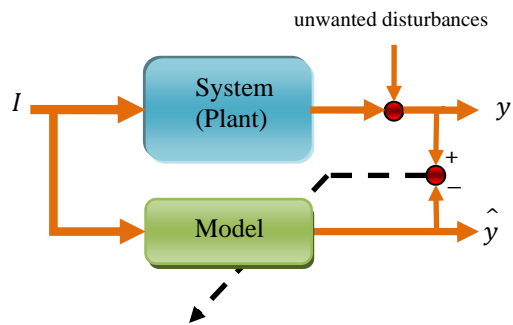

**Figure 1 General structure of modeling**

Inputs, *I*, simultaneously enters into system and model. Due to many reasons, like noises or un-calibrated sensors in industrial plants; round off errors in numerical problems; fraudulent data in social time series, etc., it's usually impossible to accurately read the true outputs of system. Instead, almost always polluted data are available. The task of a modeling is to construct an analytical tool that mimics the real system as accurate as possible and robust against unwanted disturbances. To reach this, discrepancy between model output and system's, get feed backed into model. This error helps to improve models performance and the process repeat till suitable model appears.

We assumed samples of both inputs and outputs of the system (and model) are available or it's reachable. Because of this assumption the modeling process is also named "Supervised Learning", or simply "Learning" process. Currently diverse forms of modeling are under study in literatures among them we may tip adaptive modeling, dynamic models and nonlinear dynamic systems are well established. In this note we will not consider all these, but instead modeling process in its general form has been investigated.

Instead of giving a train of algorithms, structures and methods without any sense on what's happening inside them, we prefer in concept introduction in a firm manner to keep the text plain and more meaningful. Materials arrangement is as follow: section two we will have a review over modeling and supervised learning with emphasis on key theorems and challenges; Section three is about active learning and surrogate modeling; section four describes some of the ongoing directions.

## 2. MODELLING AND SUPREVISED LEARNING

Fig.2 illustrates steps to construct a model for a system. As the steps goes down the needs to prior knowledge on the system is replaced by experiments, and vice versa.

These steps in an agile manner try to find some answers for the the following choices:

- Choice of model inputs.
- Choice of the excitation signals
- Choice of model architecture
- Choice of the model structure & complexity
- Choice of model parameters.
- Model Validation

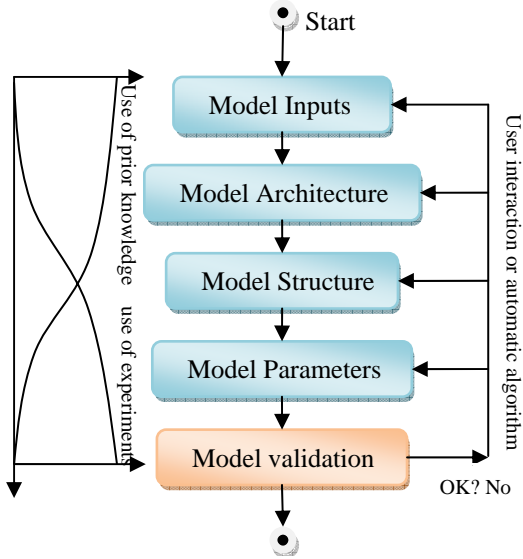

**Figure 2- Model Construction steps**

## 2.1 Model Inputs & Excitation Signals

Step 1 is typically realized on trial and error approach with the help of prior knowledge. In mechanical processes the influence of the different variables is usually quite clear, however as the process categories become more complex, e.g. chemical, biological economic, the number of potential model inputs typically increases, and the insight into the influence of different variables decreases. To solve the problem bunch of tools has been developed, among them Principle Component Analysis (PCA) and Evolutionary Algorithms (EA) has been widely applied.

Constructed model is strongly depend on the quality of inputs were used during the modeling (training). If the inputs were such that most hidden characteristics[1] of the system get revealed, then the model has the chance to learn them, otherwise model will suffer from serious inaccuracy. Despite this clear and important relation there are a few study in the realm of supervised learning has been dedicated to the topic. As we will see surrogate modeling systematically consider this need.

## 2.2 Model Architecture

Deciding on the architecture of a model depend on many facts, like: *Problem type*, *Problem dimensionality*, *resources restrictions*, *user experiments* and *customer acceptance*. Unfortunately, there is no general method which gives the best architecture for every problem. Its only some guidelines based on reported studies are available that advise for what problem types which sort of modeling architecture may suit more.

Classically, modeling tools with respect to their information sources are categorized in three main forms: "*White Box*", "*Gray Box*" and "*Black Box*". Table1 compares these categories with each other. Historically, Those believed in white box modeling confined themselves to just analytical study of system. Outputs of their works were usually in the form of heavy, expensive but precise simulators. Finite Element Analysis (FEA), Computational Fluid Dynamics (CFD) are among them. These folks are usually strongly suspicious against Black box modeling in which just a bunch of samples are all to make a model. Despite this, tendency toward black box modeling is increasing, recently. As already mentioned increasing in systems complexity and reducing in computation cost are among reasons that make this shift reasonable. In the middle, there are some whom believes in these two simultaneously. Fuzzy movement is most brilient example of this category.

**Table 1 Model types**

|  | White box | Gray box | Black Box |
|---|---|---|---|
| Information sources | First priciples insights | Insights + Data | **E**xperiments + **D**ata |
| Features | **G**ood extrapolation **G**ood understanding **H**igh reliability Scalable |  | **S**hort development time **L**ittle domain experties req. **U**sable for not understood systems |
| Drawbacks | **T**ime consuming **D**etailed domain experties req. **K**nowledge restricts accuracy **O**nly for well understood proc. |  | **N**o reliable extrapolation **N**ot scalable **D**ata restric accuracy **L**ittle understanding |
| Application area | **P**lanning, construction, design **R**ather simple process |  | **O**nly for existing process **R**ather complex processes |

## 2.3 Model Structure

This step is usually is much harder rather than next step, parameter estimation. It can be automatically carried out if structure optimization techniques such as orthogonal least square (OLS) for linear parameterized models or evolutionary algorithms (EA's) for non linear parameterized model are applied. An alternative to these general approaches is model specific growing and/or pruning algorithms.

At this step one of the most fundamental pillars of machine learning must be concidered: "*Bias/Variance*" trade off. Usually, models are simpler approximate representation of a complex system. Because of this approximation, it's quite reasonable that models outputs suffer from a *bias* from reality. Bias reduction is possible by increasing model's complexity, e.g. putting more neurons into ANN, using higher order polynomials in polynomial regressions, etc. In addition models parameters are optimized with respect to limited set of data, which by itself imposes some error in models outputs. This type of error is well known as model *variance* .To solve this problem one way is to have large amount (infinite count) of samples of system which is not the case for real problems. Other method for variance reduction is to reduce the system's complexity. So, increasing model complexity reduces model bias but simultaneously cause in model variance and vase versa. This trade off hinder automation of this step and makes it more application dependent.

---
[1] System engineers call this "System's Dynamics"

### 2.4 Model Parameters

This step is easiest to automate. Parameters of a model estimate with regards to different criterions, e.g. Entropy, Log-Likelihood, minmax, etc. choosing appropriate estimator, consistency of estimation, and type of estimated parameters (linear/nonlinear) are major issues. Currently amount of good optimization toolboxes are available which seems to be useful.

### 2.5 Model Validation

After model construction, how well it works on unseen inputs? This question is answered by Vapnic, in which he gave an upper bound for *generalization error* (models error for unseen inputs). He proved that this bound is dominated by problem complexity and total number of samples which used for model construction. This boundary may be count as another pillar of machine learning which must be considered when ones wants to judge generated model, "*no pain, no gain, give me more samples, I'll give you more accurate model!*".

Model validation is applicable in two forms. *Hold out* and *cross validation*. Hold out method is done by partitioning available samples into two categories, "*Training*", "*Validation*". Models are trained on training samples and get tested by validations. This methodology suffers from a great weak point, wasting samples. On the other hand, *Cross Validation* as a family of algorithms which more efficiently uses data has been proposed. *k-fold* cross validation is a member of this family, widely used in literature. *leave-one-out* is the one which is used for missing data prediction models.

Validation is evaluted by different measures which among them R-square, RMSE, Max are more famous.

### 3. ACTIVE LEARNING AND SURROGATE MODELING

In modeling process it's usually assumed enough amount of samples are available or it's easy to achieve. This assumption is not true always specially for complex systems. Indeed, samples are often expensive, time consuming and rare. Many systems, today have their own simulators. The disadvantage with these simulators are that they are usually expensive and sometimes slow.

Surrogate modeling[2] (in general active learning) is a supervised machine learning technique in which the learner is in control of the data used for learning. Fig.3 represents its internal loop.

Surrogate modeling may be accomplished to construct a model which mimic locally or globally the behavior of system. Local models are usually used in conjugation with optimizers which helps a faster way to find optimum point(s). On the other hand global surrogate models give a deep insight into the internal structure of a system.

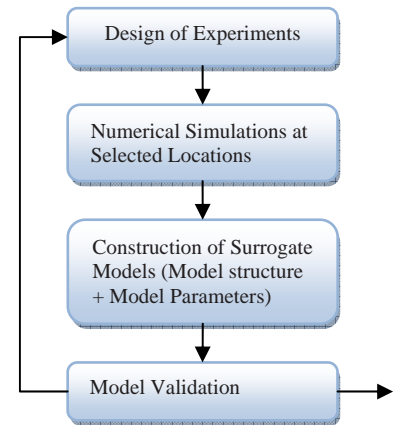

**Figure 3 surrogate modeling loop**

**Designe of experiments[3]**:

This step is an indispencible parts of surrogate modeling which there is no counter part in standard form of modeling. Originated from a theory with same name, in which the objective is to planning experiments so that the random error in physical experiments has minimum influence in the approval or disapproval of a hypothesis. As example we can name grids, Latin hypercubes, Mont Carlo and uniform designs. Generally DoE algorithms are one of two types of Exploitation or Exploration. The first one tries to choose samples such that get as much information by themselves as possible, while the second type chooses samples uniformly from design space. Recently new algorithms has been proposed such that has pros of both these types, e.g. LOLA-VERONI.

In general, more sample points offer more information of function, however, at a higher expense for low order functions, after reaching a certain sample size, increasing the number of sample points does not contribute to the approximation accuracy.

**Numerical Simulation:**

After making decision on next samples which must be evaluate these points sent into system/simulator to receive outputs.

**Construction of surrogate modeling:**

Like classical model design, for surrogate modeling there are a bunch of options available to construct a model based on, e.g. Support Vector Machines (SVM),

---

[2] surface modeling, meta-modeling

[3] In some references this step is also named sampling

Artificial Neural Networks (ANN),Kriging, Polynomial regressors, spline, Guassian Process, Radial Basis functions, etc.

As mentioned earlier, in classical modeling tools choosing appropriate model is not directly considered inside the training loop, and it's a matter that depends more on personal experiences. Inspite of this, there are some reports on surrogate modeling which automate this task.

Surrogate modeling is usually considered as black box modeling. In its pure form this claim may be correct but in available toolboxes, SUMO, iSight, etc. its also possible to guid the modeling toolbox by means of some constrains which experts propose. Due to this sorrogate modeling may be either considered as gray box modeling.

**Validation Process:**

This step is almost similar to the one that already dis cussed in the orevious section.

Surrogate modeling has been riveted many attentions all over the globe. Currently there are many free and commercial tools which its complete list and comparisons are available at [7]. These toolboxes in real world applied to many problems. These problems are usually fall into one of four types of:
**Model Approximation**, **Design Space exploration** in which visual view over system is aimed, **problem formulation** in which metamodel helps engineer to choose appropriate optimization criterion and suitable boundaries for constraints and, finally **optimization support** in which surrogate model is used as a substitute of classical gradient based optimization. Successful implementation of this last usage represented a good picture of surrogate modeling, which led to Metamodel-based Design Optimization (MBDO).

Befor end this line, we have to emphesise that Fig.3 is not the only iteration loop. For optimization problems some other loops are also available.

## 4. FUTURE WORKS

Surrogate modeling has to come with a solution to challenges like curse of dimensionality, computational complexity, noises and meta-model validation. Each of the following directions must be considered with aforementioned challenges.

### 4.1 Large scale problems

Current tendency in the realm of system identification is toward to linear sub-system approximations of high nonlinear complex systems. Successful structures like Local Linear Model Tree (LOLIMOT), TSK-fuzzy systems, etc. are some of the widely used. Adapting these structures for metamodeling is seems to be an indispensible need for future of surrogate modeling. In this context new methods for splitting models, intelligent sample selection and model concatenation are in horizon challenges

### 4.2 Domain expert knowledge

A thing that separate fuzzy systems from all available is in its systematic ability to incorporate knowledge of human expert in modeling process. For surrogate modeling this knowledge incorporation is usually appeared as a set of constraints, e.g. optimization constraints in MDO. Increasing studies in this realm may result in more generic, speedy and accurate models.

### 4.3 Uncertainity in Metamodeling

Metamodeling can be used to filter noises in computer simulation. On the other hand the uncertainty in metamodels brings new challenges in design optimization. For constrained optimization problems, if both constraint and objective functions are computation expensive and metamodeling is applied, it is found that the constrained optimum is very sensitive to the accuracy of all metamodels. Mathematically rigorous methods have to developed to qualify the uncertainty of a metamodel.